\documentclass[letterpaper]{article} 
\usepackage[preprint]{aaai2027}  
\usepackage[hyphens]{url}  
\usepackage{graphicx} 
\usepackage{natbib}  
\usepackage{caption} 
\usepackage{algorithm}
\usepackage{algorithmic}
\usepackage{amsmath,amssymb,amsfonts}
\usepackage{booktabs}
\usepackage{array}

\newcommand{\R}{\mathbb{R}}
\newcommand{\E}{\mathbb{E}}
\newcommand{\Pcal}{\mathcal{P}}
\newcommand{\Xcal}{\mathcal{X}}
\newcommand{\Zcal}{\mathcal{Z}}
\newcommand{\Lcal}{\mathcal{L}}
\newcommand{\W}{\mathsf{W}}

\newcommand{\softplus}{\operatorname{softplus}}
\newcommand{\argmin}{\operatorname*{arg\,min}}

\newcommand{\1}{\mathbf{1}}
\newcommand{\dd}{\mathrm{d}}

\graphicspath{{figures/}}
\title{Learning Predictive Ambiguity Sets for Decision-Focused Distributionally Robust Optimization}
\author{Junjie Guo}
\affiliations{
Rutgers University\\
\texttt{jg1806@scarletmail.rutgers.edu}
}

\begin{document}

\maketitle

\begin{abstract}
Predict-then-optimize systems usually compress uncertainty into a point forecast and then solve a downstream optimization problem as if the forecast were reliable. Distributionally robust optimization (DRO) offers protection against misspecification, but the ambiguity set is often centered at historical samples and uses a fixed radius. We propose \emph{learned predictive ambiguity sets} (LPAS): a deep contextual model outputs a finite nominal scenario distribution, a state-dependent Wasserstein radius, and optionally an anisotropic ground metric. These outputs define a contextual ambiguity set that feeds a DRO decision layer. The radius is trained by a combination of conditional quantile calibration, size regularization, and downstream decision loss, so that robustness is adaptive rather than globally fixed. We derive the finite dual form used by the decision layer, present a staged training algorithm, and evaluate the method on distributionally robust portfolio optimization with 20 S\&P 500 constituents from 2018--2026. The proposed method substantially improves over equal-weight, predict-then-optimize, and historical Wasserstein DRO baselines, achieving 26.28\% annualized return, Sharpe ratio 1.30, final wealth 1.61, and lower tail loss than a deep fixed-radius DRO baseline while using a smaller average radius. The results show that learned ambiguity radii can recover most of the performance of strong fixed-radius DRO while reducing unnecessary conservatism and improving regime adaptivity.
\end{abstract}

\section{Introduction}

Many machine-learning decision systems are built around the pipeline
\begin{equation}
    z_t \longrightarrow \widehat{\xi}_{t+1} \longrightarrow x_t,
\end{equation}
where $z_t$ is a context, $\widehat{\xi}_{t+1}$ is a forecast of an uncertain future quantity, and $x_t$ is a downstream decision. In portfolio optimization, $\xi_{t+1}$ represents future returns; in inventory control, it represents demand; in network optimization, it may represent edge costs. This architecture is simple, but it over-trusts the predictive model. Errors that are small in prediction space can be large in decision space, especially when the optimizer amplifies mistakes along active constraints or high-sensitivity directions.

DRO replaces a single predictive distribution by an ambiguity set and optimizes against the worst plausible distribution. A common formulation is
\begin{equation}
    \min_{x\in \Xcal} \sup_{Q\in \Pcal} \E_Q[\ell(x,\xi)],
    \label{eq:basic_dro}
\end{equation}
where $\ell$ is a downstream loss and $\Pcal$ is an ambiguity set. Wasserstein DRO is attractive because it provides a geometry-aware way to perturb an empirical or nominal distribution and often admits tractable convex reformulations \cite{mohajerin2018data,gao2023distributionally,blanchet2019quantifying,kuhn2019wasserstein}. However, the ambiguity set is commonly hand-designed: the center is a historical empirical distribution and the radius is a fixed scalar tuned by validation or statistical concentration. This can be mismatched in contextual environments. A radius that is safe in volatile periods can be too conservative in stable periods, while a radius tuned for average validation loss can fail under regime shift.

This paper asks whether the ambiguity set itself can be predicted. Given context $z_t$, a deep model outputs a finite nominal distribution
\begin{equation}
    \widehat P_\theta(\cdot\mid z_t)=\sum_{i=1}^{N}p_{\theta,i}(z_t)\delta_{\widehat\xi_{\theta,i}(z_t)},
    \label{eq:nominal_distribution_intro}
\end{equation}
plus a nonnegative radius $\rho_\phi(z_t)$. These define the contextual Wasserstein ambiguity set
\begin{equation}
    \Pcal_{\theta,\phi}(z_t)=\left\{Q:\W_c\bigl(Q,\widehat P_\theta(\cdot\mid z_t)\bigr)\le \rho_\phi(z_t)\right\}.
    \label{eq:lpas_intro}
\end{equation}
The decision is then computed by a DRO layer. The key modeling principle is that uncertainty should be both statistically calibrated and decision relevant: the radius should be large when the forecast is unreliable or the decision is sensitive to errors, and small when robustness mainly induces conservatism.

\paragraph{Contributions.}
This work makes four contributions. First, it introduces learned predictive ambiguity sets, a contextual bridge between probabilistic deep forecasting and Wasserstein DRO. Second, it develops a radius-learning objective that combines prediction loss, quantile-style calibration, radius-size regularization, and realized decision loss. Third, it gives a tractable DRO layer through the Wasserstein dual and an implementable staged training algorithm. Fourth, it provides a portfolio optimization study showing that adaptive ambiguity radii can strongly outperform nonrobust and historical-DRO baselines, and can match much of the performance of a deep fixed-radius DRO model with a smaller learned radius.

\section{Related Work}

\paragraph{Robust optimization and DRO.}
Classical robust optimization protects decisions against deterministic uncertainty sets and provides tractable reformulations for many conic and linear models \cite{benTal2009robust,bertsimas2011theory}. DRO extends this idea from uncertain parameters to uncertain probability laws. Moment-based ambiguity sets provide early data-driven DRO formulations \cite{delage2010dro,goh2010distributionally}, while modern Wasserstein ambiguity sets use optimal transport geometry to compare empirical and perturbed distributions \cite{mohajerin2018data,gao2023distributionally,blanchet2019quantifying}. Wasserstein DRO has also been connected to statistical regularization and adversarial robustness in machine learning \cite{shafieezadeh2015distributionally,sinha2018certifying,duchi2021learning,gao2022wasserstein}. LPAS keeps the tractable worst-case-expectation machinery of Wasserstein DRO, but makes the center and radius contextual and learnable.

\paragraph{Decision-focused learning.}
Smart predict-then-optimize trains prediction models by downstream decision quality rather than standard predictive error \cite{elmachtoub2022smart}. Related decision-focused approaches differentiate through combinatorial or continuous optimization layers so that the predictor is optimized for the final task \cite{wilder2019melding,amos2017optnet,agrawal2019differentiable}. LPAS follows this principle, but the learned object is not merely a point estimate or deterministic cost vector. It is a predictive distribution together with a state-dependent ambiguity radius, so the downstream layer can decide both where to optimize and how much robustness is needed.

\paragraph{Learning uncertainty sets and calibration.}
Recent work learns robust uncertainty sets from data and differentiates through robust optimization problems \cite{wang2023learning}. Predict-then-calibrate methods construct robust contextual feasible sets after fitting a predictor \cite{sun2023predict}, while end-to-end conditional robust optimization directly trains robust contextual decisions with a coverage-sensitive objective \cite{chenreddy2024end}. LPAS differs by learning a finite predictive distribution and a Wasserstein radius jointly. The radius-learning loss is also related to quantile regression and conformal calibration, which provide tools for adaptive predictive uncertainty \cite{koenker1978regression,romano2019conformalized}.

\paragraph{Portfolio optimization.}
The experimental task is rooted in mean-variance portfolio selection \cite{markowitz1952portfolio} and risk-sensitive portfolio design. Wasserstein DRO has been studied for robust mean-variance portfolios and related financial decision problems \cite{blanchet2018distributionally}. Our portfolio layer uses the predictive scenario model to estimate returns, a Wasserstein ambiguity penalty to temper aggressive forecasts, and standard turnover and covariance regularization. Tail metrics such as CVaR are reported because average returns alone can hide downside risk \cite{rockafellar2000optimization}.

\section{Problem Setup}

Let $(z,\xi)\sim P^\star$, where $z\in\Zcal$ is the observed context and $\xi\in\Xi\subseteq\R^d$ is an uncertain quantity realized after the decision. A decision maker selects $x\in\Xcal\subseteq\R^m$ and incurs loss $\ell(x,\xi)$. The goal is to learn a contextual decision rule $z\mapsto x(z)$ with low out-of-sample loss and stable tail behavior.

The proposed architecture has four components: a nominal scenario model $\widehat P_\theta(\cdot\mid z)$, a radius network $\rho_\phi(z)$, an optional ground metric $c_\psi(\xi,\xi';z)$, and a DRO decision layer. The nominal model returns scenarios $\widehat\xi_{\theta,i}(z)$ and probabilities $p_{\theta,i}(z)$ with $\sum_i p_{\theta,i}(z)=1$. The radius network is parameterized as
\begin{equation}
    \rho_\phi(z)=\rho_{\min}+\softplus(g_\phi(z)),
    \label{eq:softplus_radius}
\end{equation}
which ensures nonnegativity. A bounded alternative is
\begin{equation}
    \rho_\phi(z)=\rho_{\min}+ (\rho_{\max}-\rho_{\min})\sigma(g_\phi(z)).
    \label{eq:bounded_radius}
\end{equation}
The optional metric may be fixed, such as $c(\xi,\xi')=\|\xi-\xi'\|_2$, or contextual and anisotropic, such as
\begin{equation}
    c_\psi(\xi,\xi';z)=\|A_\psi(z)(\xi-\xi')\|_2.
    \label{eq:learned_metric}
\end{equation}
The experiments in this paper use a fixed Euclidean transport geometry; the learned-metric extension is left as a modular component.

\section{Learned Predictive Ambiguity Sets}

\subsection{Discrete Nominal Distributions}

For each context $z_t$, the model outputs
\begin{equation}
    \widehat P_{\theta,t}=\sum_{i=1}^{N}p_{\theta,i}(z_t)\delta_{\widehat\xi_{\theta,i}(z_t)}.
    \label{eq:nominal_distribution}
\end{equation}
The predictive mean is
\begin{equation}
    \widehat\mu_{\theta,t}=\E_{\widehat P_{\theta,t}}[\xi]=\sum_{i=1}^{N}p_{\theta,i}(z_t)\widehat\xi_{\theta,i}(z_t).
    \label{eq:predictive_mean}
\end{equation}
A point-prediction baseline uses only $\widehat\mu_{\theta,t}$, while LPAS uses the full finite distribution as the center of a Wasserstein ball.

\subsection{Wasserstein Ambiguity Set}

For a distribution $Q\in\Pcal(\Xi)$ and nominal distribution $\widehat P_{\theta,t}$, define
\begin{equation}
    \W_{c_\psi}(Q,\widehat P_{\theta,t};z_t)=
    \inf_{\gamma\in\Pi(Q,\widehat P_{\theta,t})}
    \int_{\Xi\times\Xi}c_\psi(\xi,\widehat\xi;z_t)\dd\gamma(\xi,\widehat\xi),
    \label{eq:wasserstein_general}
\end{equation}
where $\Pi(Q,\widehat P_{\theta,t})$ is the set of couplings with marginals $Q$ and $\widehat P_{\theta,t}$. The learned ambiguity set is
\begin{equation}
    \Pcal_{\theta,\phi,\psi,t}=\{Q\in\Pcal(\Xi):\W_{c_\psi}(Q,\widehat P_{\theta,t};z_t)\le \rho_\phi(z_t)\}.
    \label{eq:learned_ambiguity_set}
\end{equation}
If $Q=\sum_{j=1}^{M}q_j\delta_{\xi_j}$ is also discrete, then the transport discrepancy is the linear program
\begin{equation}
    \min_{\pi\ge 0}\sum_{j=1}^{M}\sum_{i=1}^{N}c_\psi(\xi_j,\widehat\xi_{\theta,i};z_t)\pi_{ji}
    \label{eq:discrete_transport_lp}
\end{equation}
subject to $\sum_j\pi_{ji}=p_{\theta,i}(z_t)$ and $\sum_i\pi_{ji}=q_j$. Thus the radius controls the amount of adversarial mass transportation away from the predicted scenarios.

\subsection{DRO Decision Layer}

Given the ambiguity set, the decision is
\begin{equation}
    x_t^\star=\argmin_{x\in\Xcal}\left\{\sup_{Q\in\Pcal_{\theta,\phi,\psi,t}}\E_Q[\ell(x,\xi)]+r(x)\right\},
    \label{eq:dro_decision_layer}
\end{equation}
where $r(x)$ is a deterministic regularizer such as a risk penalty, transaction-cost term, or strong-convexity penalty.

The standard Wasserstein dual gives
\begin{equation}
\begin{aligned}
    &\sup_{Q:\W_{c_\psi}(Q,\widehat P_{\theta,t})\le \rho_t}
    \E_Q[\ell(x,\xi)] \\
    &\quad =\inf_{\eta\ge 0}\left\{\eta\rho_t+
    \sum_{i=1}^{N}p_{\theta,i}(z_t)\,s_i(x,\eta)\right\},
\end{aligned}
\label{eq:wasserstein_dual}
\end{equation}
where $\rho_t=\rho_\phi(z_t)$ and
\begin{equation}
    s_i(x,\eta)=\sup_{\xi\in\Xi}
    \bigl[\ell(x,\xi)-\eta c_\psi(\xi,\widehat\xi_{\theta,i};z_t)\bigr].
\end{equation}
Under affine loss and norm-based costs, the inner supremum has a finite-dimensional conic representation or a closed form. This is the computational bridge from learned ambiguity sets to differentiable robust decisions.

\subsection{Portfolio Special Case}
\label{sec:portfolio_special_case}

The portfolio task provides a useful closed-form instance of the decision layer. Let $r\in\R^d$ be next-period returns and let $w\in\Delta_d$ be long-only portfolio weights,
\begin{equation}
    \Delta_d=\{w\in\R_+^d:\1^\top w=1\}.
\end{equation}
A transaction-cost and risk-regularized loss is
\begin{equation}
    \ell(w,r)=-r^\top w+\lambda_{\mathrm{risk}}w^\top\widehat\Sigma_t w+
    \lambda_{\mathrm{tc}}\|w-w_{t-1}\|_1.
    \label{eq:portfolio_loss}
\end{equation}
The corresponding DRO decision is
\begin{equation}
\begin{aligned}
    w_t^\star=\argmin_{w\in\Delta_d}\quad &
    \sup_{Q:\W_c(Q,\widehat P_{\theta,t})\le \rho_\phi(z_t)}\E_Q[-r^\top w] \\
    &+\lambda_{\mathrm{risk}}w^\top\widehat\Sigma_t w+
    \lambda_{\mathrm{tc}}\|w-w_{t-1}\|_1.
\end{aligned}
\label{eq:portfolio_dro}
\end{equation}
If $c(r,\widehat r)=\|r-\widehat r\|_2$ and only the linear return term is robustified, then duality gives
\begin{equation}
    \sup_{Q:\W_1(Q,\widehat P_{\theta,t})\le \rho_t}\E_Q[-r^\top w]
    =-\widehat\mu_{\theta,t}^\top w+\rho_t\|w\|_2.
    \label{eq:portfolio_closed_form}
\end{equation}
Thus the robust portfolio can be written as
\begin{equation}
\begin{aligned}
    \max_{w\in\Delta_d}\quad &\widehat\mu_{\theta,t}^\top w
    -\rho_\phi(z_t)\|w\|_2
    -\lambda_{\mathrm{risk}}w^\top\widehat\Sigma_t w \\
    &-\lambda_{\mathrm{tc}}\|w-w_{t-1}\|_1.
\end{aligned}
\label{eq:portfolio_intuitive}
\end{equation}
The radius has a direct interpretation: larger $\rho_\phi(z_t)$ makes the optimizer less aggressive when the predictive distribution is unreliable.

\subsection{Radius Calibration}

The radius should not be merely large; it should be calibrated. Let
\begin{equation}
    e_t=\left\|S_{\theta,t}^{-1/2}(\xi_{t+1}-\widehat\mu_{\theta,t})\right\|_2,
    \label{eq:normalized_error}
\end{equation}
where $S_{\theta,t}$ is a predicted or empirical regularized scale matrix. A statistically interpretable radius approximates a conditional high quantile,
\begin{equation}
    \rho_\phi(z_t)\approx Q_\tau(e_t\mid z_t),
    \label{eq:conditional_quantile_radius}
\end{equation}
with quantile level $\tau$. We use the pinball loss
\begin{equation}
    \ell_\tau(e,\rho)=(\tau-\1\{e\le \rho\})(e-\rho).
    \label{eq:pinball_loss}
\end{equation}
This gives the learned radius a forecast-calibration interpretation before it is fine-tuned for downstream decision quality.

\section{Training Objective and Algorithm}

The complete objective is
\begin{equation}
\begin{aligned}
    \min_{\theta,\phi,\psi}\quad
    \Lcal(\theta,\phi,\psi)&=\Lcal_{\mathrm{dec}}+
    \lambda_{\mathrm{pred}}\Lcal_{\mathrm{pred}}+
    \lambda_{\mathrm{cal}}\Lcal_{\mathrm{cal}} \\
    &\quad +\lambda_{\mathrm{size}}\Lcal_{\mathrm{size}}+
    \lambda_{\mathrm{stab}}\Lcal_{\mathrm{stab}}.
\end{aligned}
\label{eq:training_objective}
\end{equation}
The decision loss is
\begin{equation}
    \Lcal_{\mathrm{dec}}=\frac{1}{T}\sum_{t=1}^{T}\ell_{\mathrm{eval}}(x_t^\star,\xi_{t+1}).
\end{equation}
The prediction loss trains the nominal scenario distribution. Depending on the scenario model, it can be a negative log-likelihood, energy score, or scenario reconstruction loss
\begin{equation}
    \Lcal_{\mathrm{pred}}=\frac{1}{T}\sum_{t=1}^{T}\min_i\|\xi_{t+1}-\widehat\xi_{\theta,i}(z_t)\|_2^2.
\end{equation}
The calibration, size, and stability losses are
\begin{equation}
    \Lcal_{\mathrm{cal}}=\frac{1}{T}\sum_{t=1}^{T}\ell_\tau(e_t,\rho_\phi(z_t)),\quad
    \Lcal_{\mathrm{size}}=\frac{1}{T}\sum_{t=1}^{T}\rho_\phi(z_t),
\end{equation}
\begin{equation}
    \Lcal_{\mathrm{stab}}=\frac{1}{T-1}\sum_{t=2}^{T}|\rho_\phi(z_t)-\rho_\phi(z_{t-1})|^2.
\end{equation}
The size term prevents radius inflation, while the calibration term prevents radius collapse. Algorithm~\ref{alg:lpas} summarizes the training procedure.

\begin{algorithm}[tb]
\caption{Training Learned Predictive Ambiguity Sets}
\label{alg:lpas}
\begin{algorithmic}[1]
\REQUIRE Context-outcome pairs $\{(z_t,\xi_{t+1})\}_{t=1}^{T}$, scenario count $N$, quantile level $\tau$
\STATE Pretrain $\widehat P_\theta(\cdot\mid z)$ by $\Lcal_{\mathrm{pred}}$
\STATE Compute normalized forecast errors $e_t$ on a calibration split
\STATE Train $\rho_\phi(z)$ by the pinball loss $\Lcal_{\mathrm{cal}}$ with size regularization
\FOR{each training epoch}
    \FOR{each minibatch of contexts}
        \STATE Output scenarios $\widehat\xi_{\theta,i}(z_t)$, probabilities $p_{\theta,i}(z_t)$, and radius $\rho_\phi(z_t)$
        \STATE Construct $\Pcal_{\theta,\phi,\psi,t}$ by Eq.~\eqref{eq:learned_ambiguity_set}
        \STATE Solve the DRO layer Eq.~\eqref{eq:dro_decision_layer}
        \STATE Evaluate realized decision loss and update parameters using Eq.~\eqref{eq:training_objective}
    \ENDFOR
\ENDFOR
\RETURN Trained scenario generator, radius network, and DRO decision rule
\end{algorithmic}
\end{algorithm}

\section{Experiments}

\subsection{Portfolio Optimization Setup}

We evaluate LPAS on daily portfolio optimization using 20 S\&P 500 constituents over January 2018--June 2026. The experiment uses daily returns from TMUS, PPG, LYB, ALB, GPC, STE, WRB, VRTX, BLK, VEEV, INCY, PAYX, TXT, CSCO, MOS, OXY, HON, MAS, AMGN, and IEX. After feature construction, the sample contains 2054 usable feature rows. The chronological split has 1129 training observations, 410 validation observations, and 515 test observations. The portfolio is long-only and fully invested.

The predictive model is a Transformer scenario generator \cite{vaswani2017attention} with lookback length 63, two Transformer layers, four attention heads, hidden dimension 48, and $N=7$ scenarios. Lagged returns, rolling volatility, and market-state features are used as contexts. The selected candidate uses ridge coefficient 1.0, risk penalty 8.0, transaction-cost objective weight 0.0015, realized transaction cost 0.001, quantile level $\tau=0.9$, and radius floor $\rho_{\min}=10^{-4}$.

The baselines are equal weight (EW), predict-then-optimize (P2O) using the predictive mean, historical Wasserstein DRO (Hist-WDRO) centered at historical returns, and deep predictive fixed-radius DRO (Fixed-DRO) centered at the same Transformer scenario distribution but using a scalar validation-selected radius. LPAS-W uses the same scenario generator as Fixed-DRO but replaces the global radius with a learned context-dependent radius.

\subsection{Hyperparameter Selection}

Hyperparameters are selected on the validation window by a decision-aware score that combines average loss, tail loss, calibration, and radius size. Table~\ref{tab:hparam} shows the candidate-level validation results. The selected model does not minimize prediction MSE alone; it minimizes the downstream decision-aware validation score.

\begin{table}[t]
\centering
\scriptsize
\setlength{\tabcolsep}{3.2pt}
\resizebox{\columnwidth}{!}{%
\begin{tabular}{lrrrr}
\toprule
Candidate & Val. Score & Decision & MSE & Selected \\
\midrule
0 & 0.000922 & 0.000909 & 0.000676 & No \\
1 & 0.001562 & 0.001549 & 0.000674 & No \\
2 & \textbf{0.000800} & \textbf{0.000784} & 0.000772 & Yes \\
3 & 0.001239 & 0.001226 & 0.000652 & No \\
4 & 0.001282 & 0.001269 & \textbf{0.000633} & No \\
\bottomrule
\end{tabular}%
}
\caption{Transformer hyperparameter selection. The selected model minimizes the decision-aware validation score rather than prediction MSE alone.}
\label{tab:hparam}
\end{table}

\subsection{Main Results}

Table~\ref{tab:main_results} reports the main out-of-sample portfolio metrics. LPAS strongly improves over equal weight, predict-then-optimize, and historical Wasserstein DRO. Compared with deep fixed-radius DRO, LPAS gives slightly lower annualized return and Sharpe ratio, but uses a substantially smaller average radius and obtains slightly better worst-month and tail-loss metrics. This suggests that the learned radius reduces conservatism while retaining most of the performance of the best fixed-radius robust model.

\begin{table*}[t]
\centering
\scriptsize
\setlength{\tabcolsep}{3.0pt}
\resizebox{\textwidth}{!}{%
\begin{tabular}{lrrrrrrrrr}
\toprule
Method & Ann. Ret. & Ann. Vol. & Sharpe & Max DD & Turnover & Wealth & Worst Mo. & CVaR95 & Radius \\
\midrule
EW & 0.0690 & 0.1622 & 0.4925 & -0.1867 & 0.0000 & 1.1462 & -0.0726 & 0.0218 & -- \\
P2O & -0.2517 & 0.3231 & -0.7365 & -0.4728 & 0.2869 & 0.5530 & -0.1736 & 0.0474 & -- \\
Hist-WDRO & 0.0455 & 0.1484 & 0.3737 & -0.1308 & 0.0659 & 1.0951 & -0.0597 & 0.0212 & 7.9481 \\
Fixed-DRO & \textbf{0.2748} & 0.1950 & \textbf{1.3428} & -0.1490 & 0.1595 & \textbf{1.6423} & -0.0714 & 0.0266 & 35.4438 \\
LPAS-W & 0.2628 & 0.1937 & 1.3019 & -0.1495 & 0.1672 & 1.6110 & -0.0703 & 0.0264 & 24.3466 \\
\bottomrule
\end{tabular}%
}
\caption{Main out-of-sample portfolio results. EW: equal weight; P2O: predict-then-optimize; Hist-WDRO: historical Wasserstein DRO; Fixed-DRO: deep predictive fixed-radius DRO; LPAS-W: learned predictive Wasserstein ambiguity set. Higher annualized return, Sharpe ratio, and wealth are better; lower maximum drawdown, turnover, worst-month loss magnitude, CVaR95 loss, and radius are better.}
\label{tab:main_results}
\end{table*}

\begin{figure}[t]
\centering
\includegraphics[width=0.98\columnwidth]{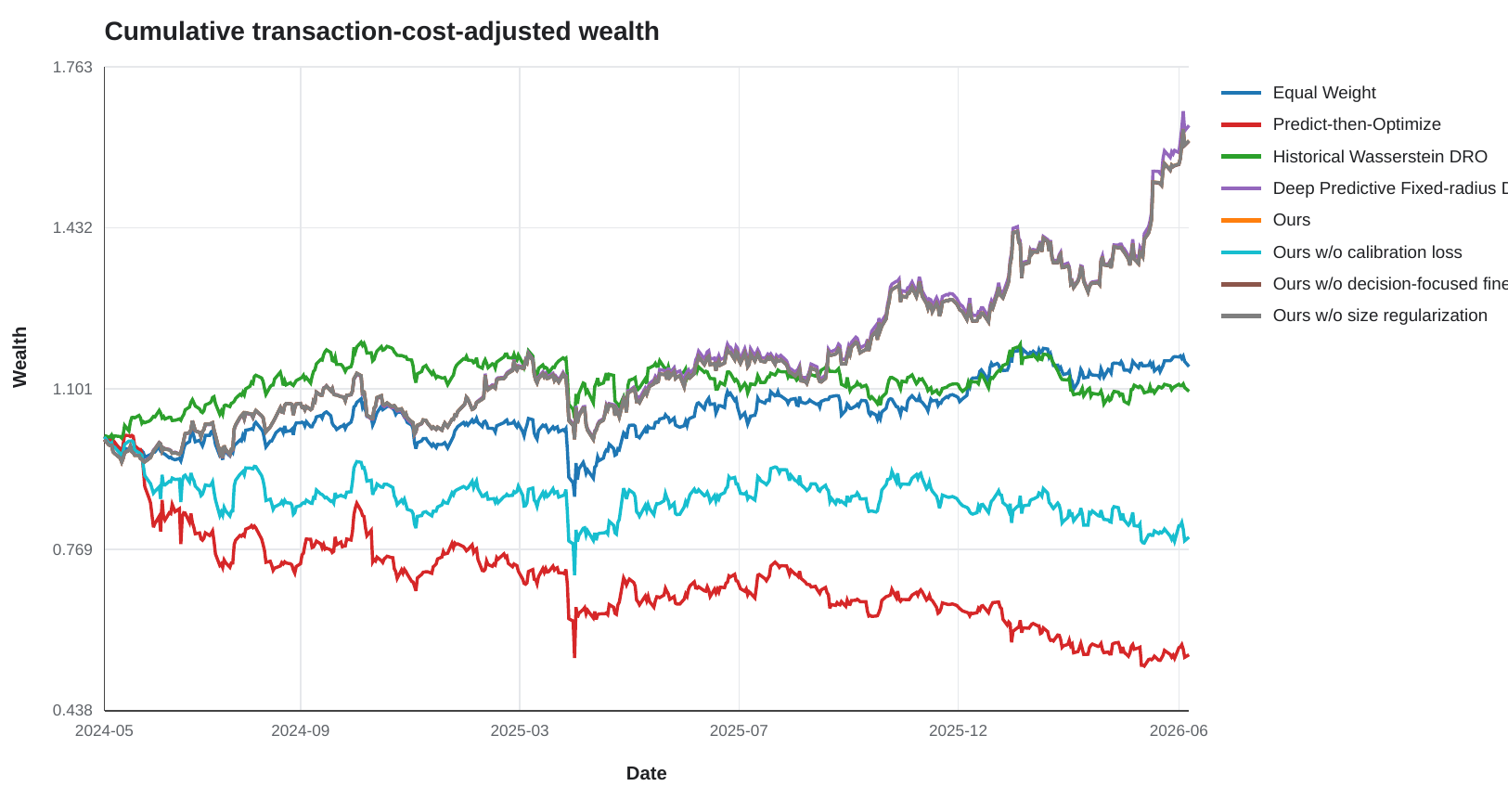}
\caption{Out-of-sample cumulative wealth. Robust methods stabilize wealth accumulation relative to predict-then-optimize.}
\label{fig:cumulative_wealth}
\end{figure}

\begin{figure}[t]
\centering
\includegraphics[width=0.98\columnwidth]{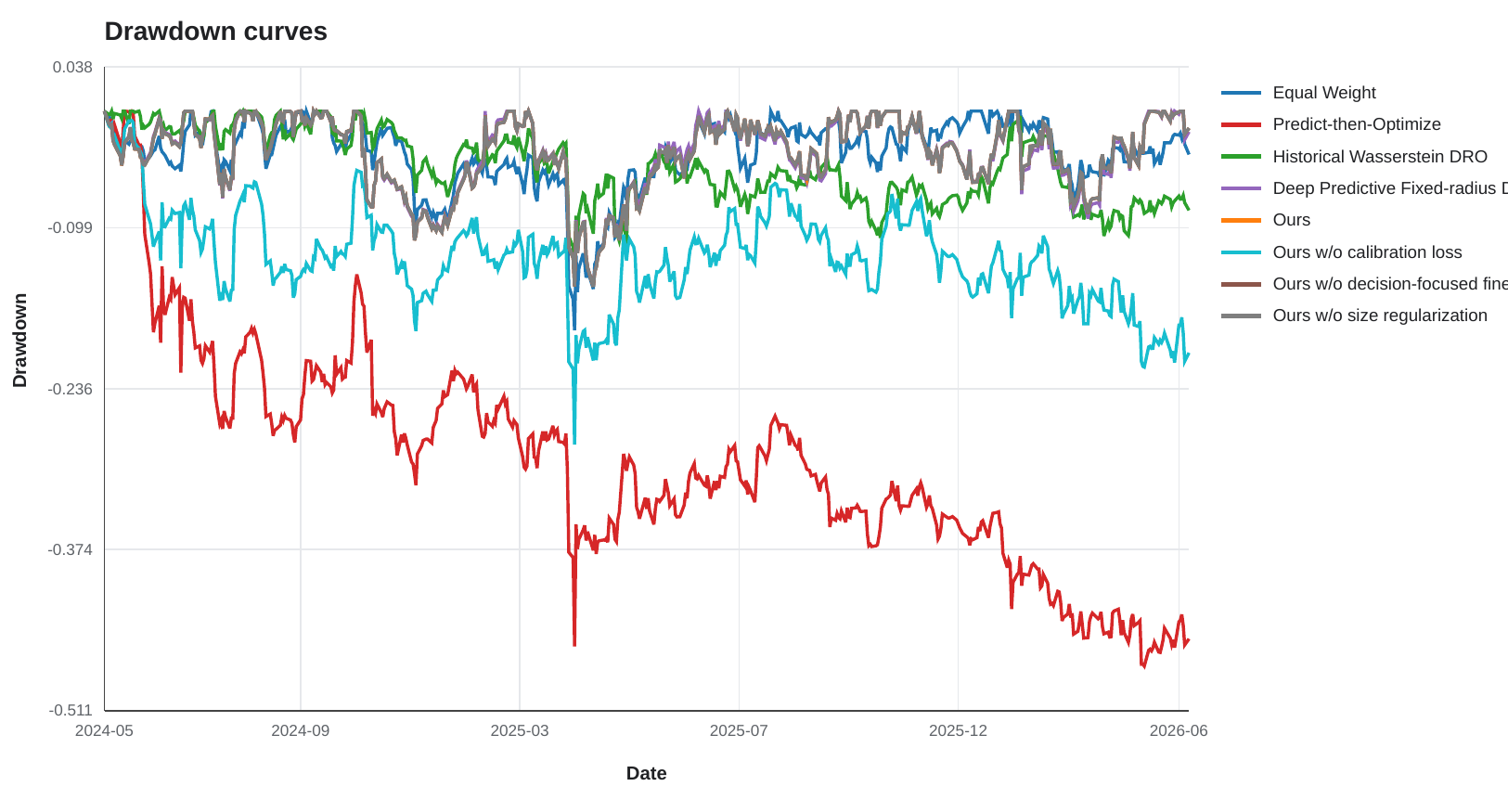}
\caption{Out-of-sample drawdown. LPAS-W remains close to the strong fixed-radius DRO baseline while avoiding a globally large ambiguity radius.}
\label{fig:drawdown}
\end{figure}

Figures~\ref{fig:cumulative_wealth} and~\ref{fig:drawdown} show cumulative wealth and drawdown separately. Predict-then-optimize suffers from aggressive forecast amplification, while robust methods produce more stable wealth paths. The learned-radius method tracks the fixed-radius robust portfolio closely while avoiding a globally large ambiguity radius.

\subsection{Ablation Results}

Table~\ref{tab:ablation} isolates the role of the learned radius components. Removing calibration collapses empirical coverage to zero and leads to negative annualized return, large drawdown, and high turnover. Removing decision-focused fine-tuning has only a small effect in this implementation, suggesting that prediction pretraining and radius calibration already explain most of the gain. Removing size regularization has little effect at the selected hyperparameter scale, but it remains important as a safeguard against radius inflation.

\begin{table*}[t]
\centering
\scriptsize
\setlength{\tabcolsep}{3.0pt}
\resizebox{\textwidth}{!}{%
\begin{tabular}{lrrrrrrrr}
\toprule
Method & Ann. Ret. & Sharpe & Max DD & Turnover & Wealth & CVaR95 & Coverage & Radius \\
\midrule
LPAS-W (Ours) & 0.2628 & \textbf{1.3019} & -0.1495 & \textbf{0.1672} & \textbf{1.6110} & \textbf{0.0264} & 0.7553 & 24.3466 \\
Without calibration loss & -0.1062 & -0.3200 & -0.2841 & 0.2626 & 0.7950 & 0.0353 & 0.0000 & 1.0000 \\
Without decision-focused fine-tuning & 0.2622 & 1.2996 & -0.1494 & 0.1676 & 1.6095 & \textbf{0.0264} & 0.7495 & 24.2486 \\
Without size regularization & \textbf{0.2628} & 1.3017 & -0.1494 & 0.1675 & 1.6109 & \textbf{0.0264} & 0.7534 & 24.3564 \\
\bottomrule
\end{tabular}%
}
\caption{Ablation study for the learned radius. Calibration is essential: without it, the learned radius collapses and robust decisions degrade.}
\label{tab:ablation}
\end{table*}

\subsection{Adaptive Radius and Calibration}

\begin{figure}[t]
\centering
\includegraphics[width=0.98\columnwidth]{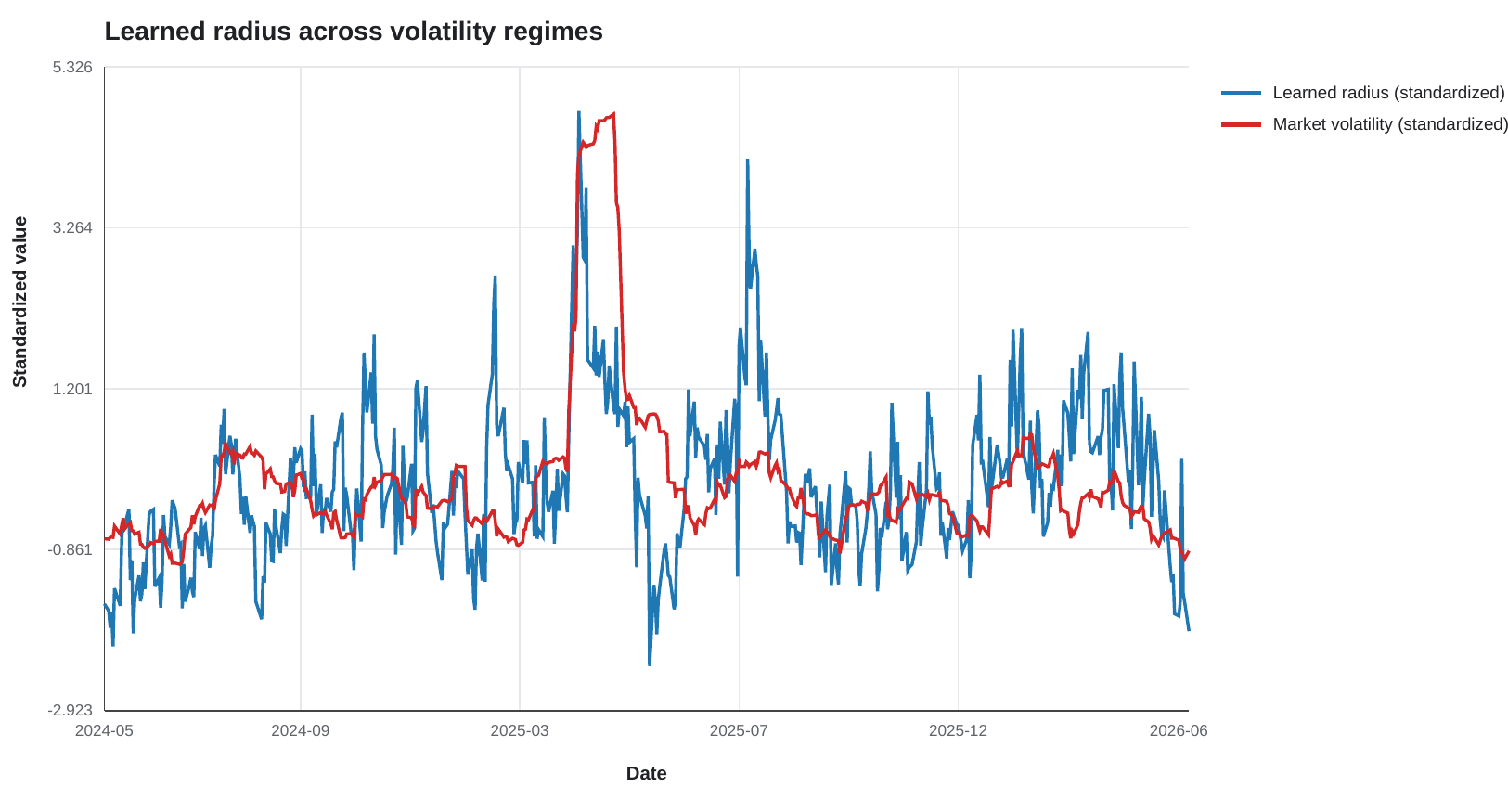}
\caption{Learned LPAS-W radius across market-volatility regimes. The radius increases in uncertain periods and decreases in calmer periods.}
\label{fig:radius_adaptivity}
\end{figure}

Figure~\ref{fig:radius_adaptivity} visualizes the learned radius against market volatility. The learned radius rises in high-uncertainty regimes and falls in calmer regimes, which is the desired behavior of a contextual ambiguity set. The empirical coverage of LPAS-W is lower than the nominal quantile target because decision-aware validation and size regularization trade exact coverage for performance and a smaller robust radius. If strict coverage is required, the radius can be post-adjusted by split conformal calibration.

\subsection{Regime Analysis}

Table~\ref{tab:regime} reports a compact regime comparison for the main methods. Each regime contains 129 test observations. LPAS-W is particularly competitive in drawdown and high learned-radius regimes. In drawdown regimes it obtains annualized return 0.5167 and Sharpe ratio 2.0312, slightly above deep fixed-radius DRO. In high learned-radius periods it also improves over the fixed-radius robust baseline in annualized return and Sharpe ratio. These results support the central hypothesis: adaptive radii are most valuable when uncertainty is state dependent and decision sensitive.

\begin{table}[t]
\centering
\tiny
\setlength{\tabcolsep}{1.3pt}
\renewcommand{\arraystretch}{0.82}
\resizebox{\columnwidth}{!}{%
\begin{tabular}{llrrrrr}
\toprule
Reg. & Method & Ret. & Sharpe & Loss & CVaR & $\rho$ \\
\midrule
Low-vol & EW & -0.0984 & -0.7086 & 0.0004 & 0.0186 & -- \\
Low-vol & Hist & -0.0516 & -0.3746 & 0.0002 & 0.0173 & 7.9481 \\
Low-vol & Fixed & \textbf{0.4405} & \textbf{2.0824} & -0.0015 & 0.0209 & 35.4438 \\
Low-vol & LPAS & 0.4094 & 2.0199 & -0.0014 & 0.0203 & 23.4948 \\
\midrule
High-vol & EW & 0.1834 & 0.8519 & -0.0008 & \textbf{0.0322} & -- \\
High-vol & Hist & 0.1770 & 0.8938 & -0.0007 & 0.0319 & 7.9481 \\
High-vol & Fixed & 0.2237 & 0.9418 & -0.0009 & 0.0351 & 35.4438 \\
High-vol & LPAS & \textbf{0.2266} & \textbf{0.9516} & -0.0009 & 0.0350 & 25.3148 \\
\midrule
Drawdown & EW & 0.4598 & 1.8638 & -0.0016 & \textbf{0.0287} & -- \\
Drawdown & Hist & 0.2477 & 1.3379 & -0.0009 & 0.0240 & 7.9481 \\
Drawdown & Fixed & 0.5103 & 2.0165 & -0.0017 & 0.0295 & 35.4438 \\
Drawdown & LPAS & \textbf{0.5167} & \textbf{2.0312} & -0.0017 & 0.0294 & 25.0710 \\
\midrule
Recovery & EW & -0.0574 & -0.3608 & 0.0002 & \textbf{0.0172} & -- \\
Recovery & Hist & 0.0460 & 0.3744 & -0.0002 & 0.0187 & 7.9481 \\
Recovery & Fixed & 0.0698 & 0.4377 & -0.0003 & 0.0251 & 35.4438 \\
Recovery & LPAS & \textbf{0.0718} & \textbf{0.4480} & -0.0004 & 0.0250 & 24.1373 \\
\midrule
Low-$\rho$ & EW & 0.0239 & 0.2455 & -0.0001 & \textbf{0.0172} & -- \\
Low-$\rho$ & Hist & 0.0909 & 0.7862 & -0.0004 & 0.0158 & 7.9481 \\
Low-$\rho$ & Fixed & \textbf{0.2680} & \textbf{1.4939} & -0.0010 & 0.0199 & 35.4438 \\
Low-$\rho$ & LPAS & 0.2324 & 1.3435 & -0.0009 & 0.0198 & 22.4219 \\
\midrule
High-$\rho$ & EW & \textbf{0.4691} & \textbf{1.8091} & -0.0016 & \textbf{0.0280} & -- \\
High-$\rho$ & Hist & 0.1530 & 0.8047 & -0.0006 & 0.0295 & 7.9481 \\
High-$\rho$ & Fixed & 0.2850 & 1.1416 & -0.0011 & 0.0363 & 35.4438 \\
High-$\rho$ & LPAS & 0.3022 & 1.1977 & -0.0012 & 0.0360 & 26.5797 \\
\bottomrule
\end{tabular}%
}
\caption{Regime results on the test set. Hist denotes historical WDRO, Fixed denotes fixed-radius DRO, and LPAS denotes LPAS-W. Bold indicates the best value within each regime.}
\label{tab:regime}
\end{table}

\begin{figure}[t]
\centering
\includegraphics[width=0.98\columnwidth]{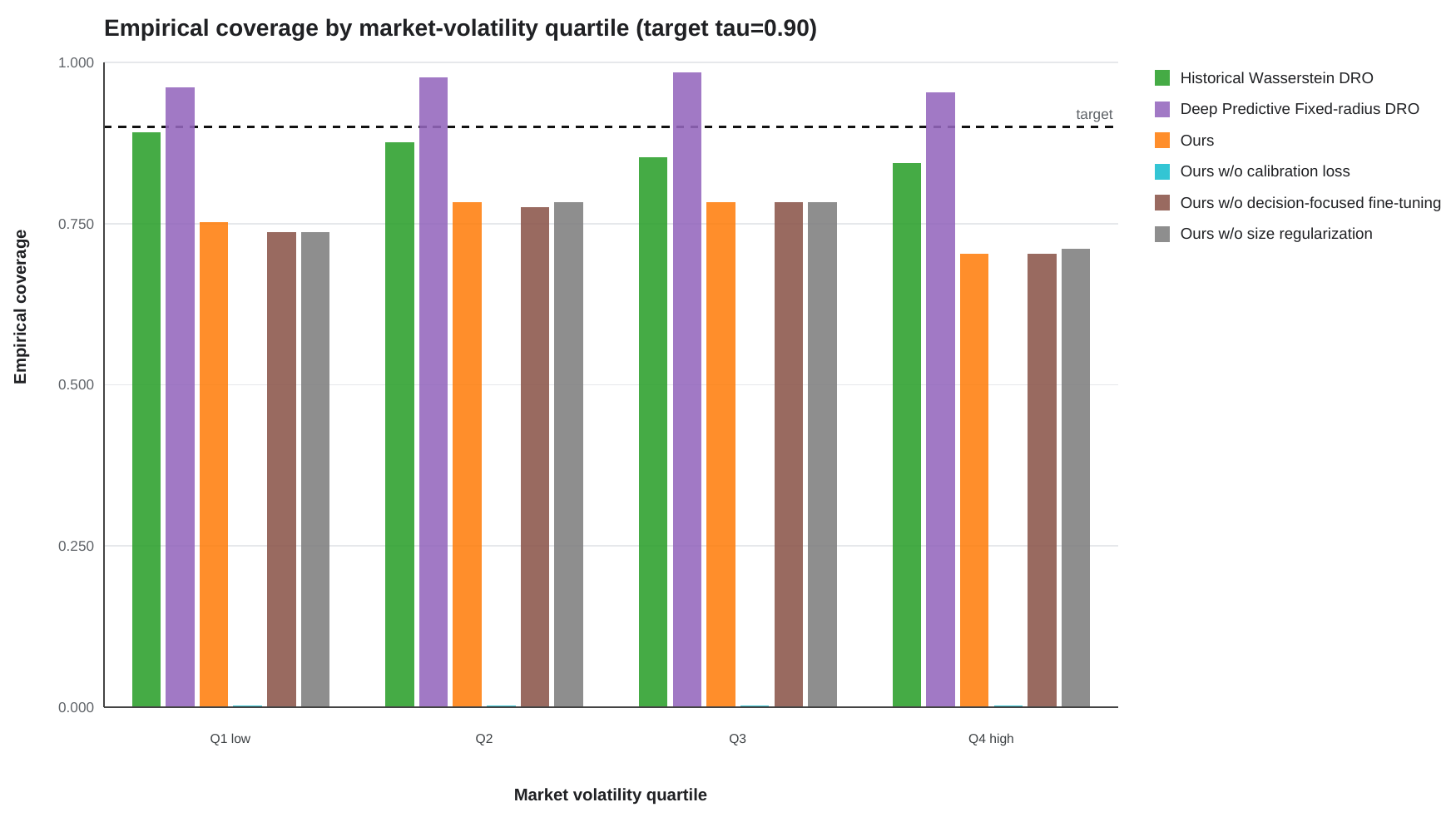}
\caption{Empirical coverage by market-volatility quartile. The diagnostic compares LPAS-W and ablations against historical and fixed-radius DRO baselines, with the dashed line marking the target coverage level.}
\label{fig:calibration_by_vol}
\end{figure}

\begin{figure}[t]
\centering
\includegraphics[width=0.98\columnwidth]{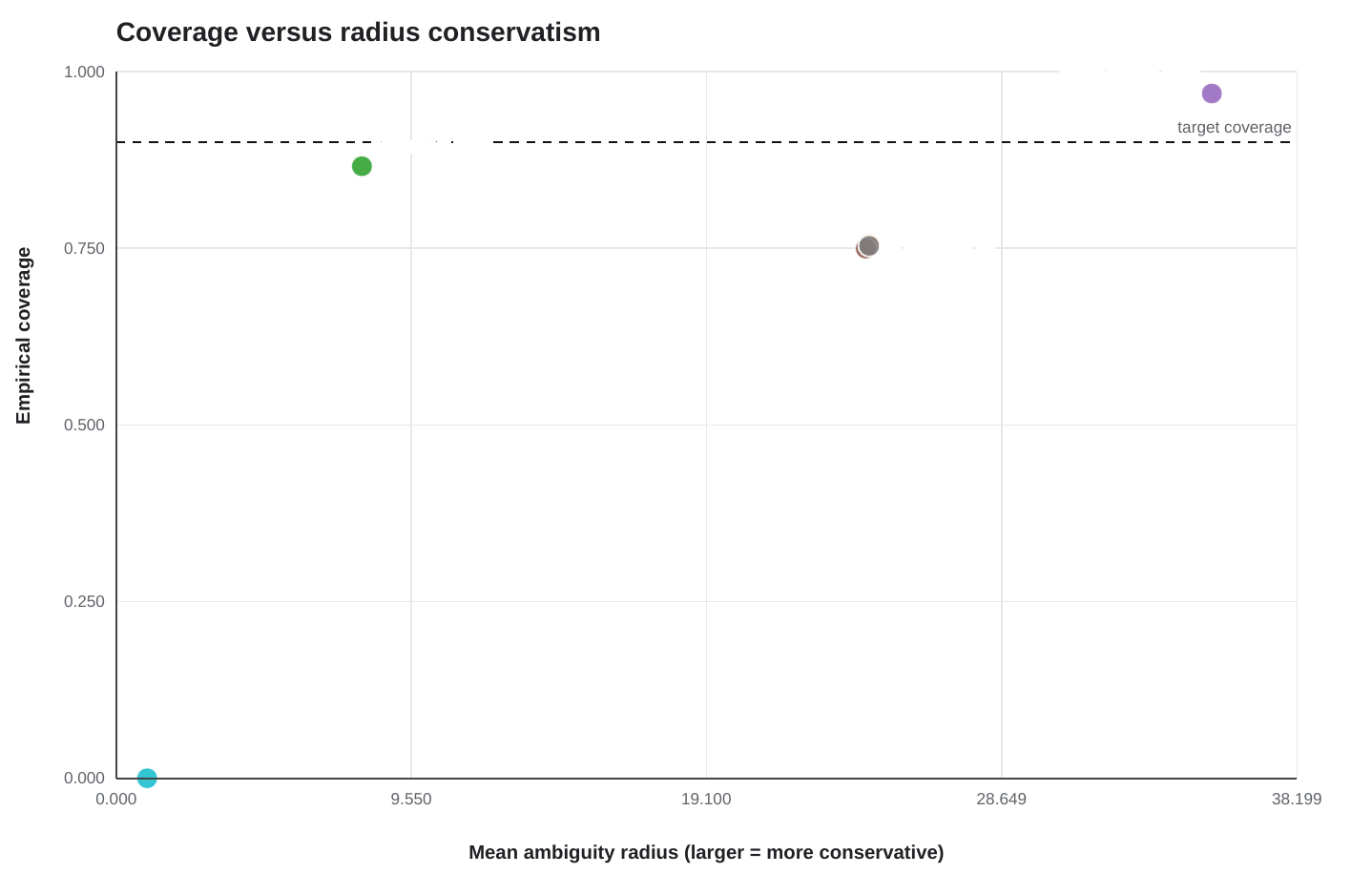}
\caption{Coverage versus radius conservatism. LPAS-W attains high empirical coverage with a smaller mean ambiguity radius than fixed-radius robust optimization, illustrating the benefit of state-dependent radius learning.}
\label{fig:calibration_conservatism}
\end{figure}

\begin{figure}[t]
\centering
\includegraphics[width=0.98\columnwidth]{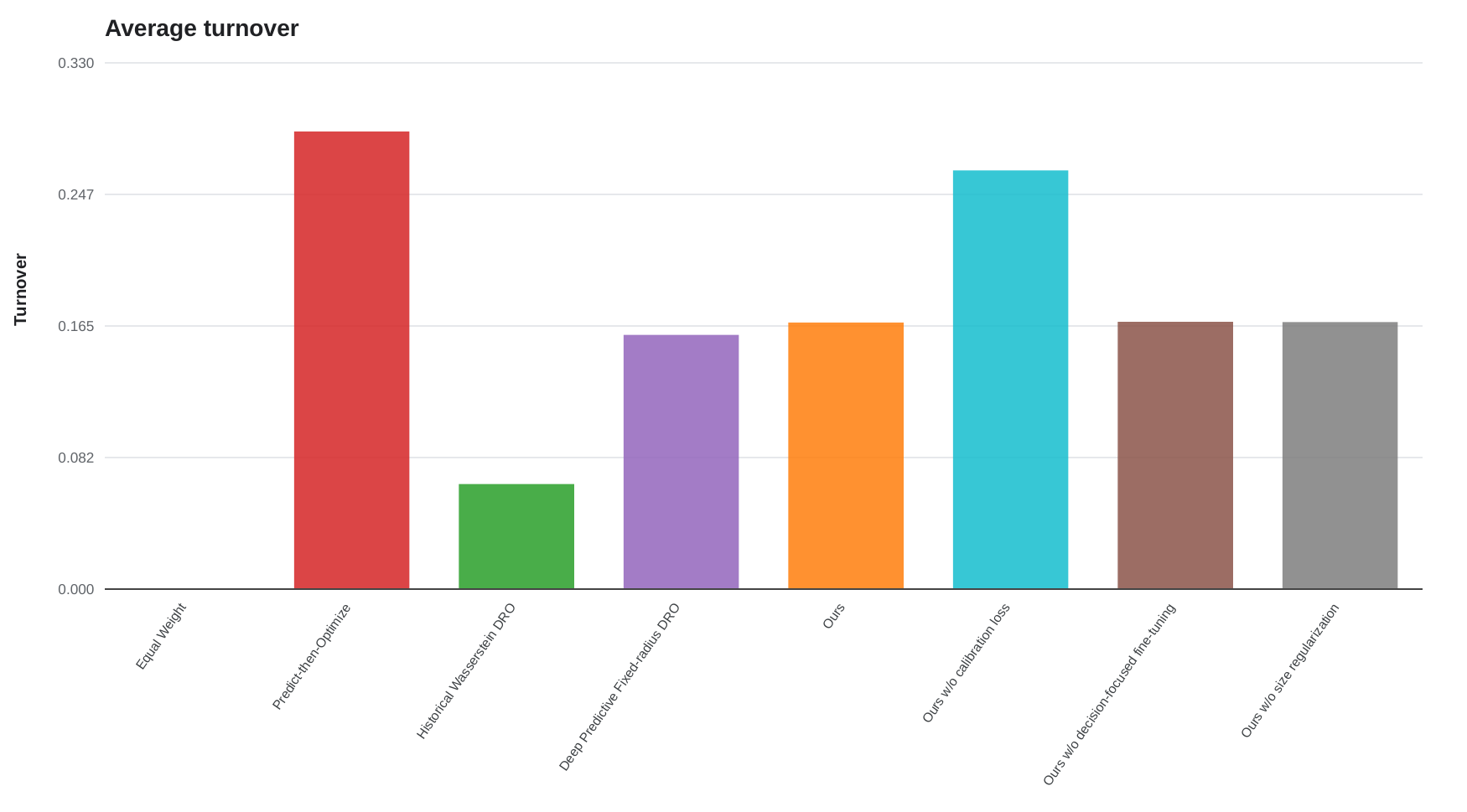}
\caption{Average turnover by method. The comparison checks whether LPAS-W's performance is driven by uncontrolled trading intensity.}
\label{fig:turnover_bar}
\end{figure}

Figures~\ref{fig:calibration_by_vol},~\ref{fig:calibration_conservatism}, and~\ref{fig:turnover_bar} give additional diagnostics. Calibration by volatility bin checks whether the learned radius responds to market state rather than acting as a constant penalty. The coverage-radius plot shows the conservatism required by each method to obtain empirical coverage, and the turnover plot verifies that the performance gain is not simply produced by uncontrolled trading intensity.

\subsection{Decision Sensitivity Diagnostic}

\begin{figure}[t]
\centering
\includegraphics[width=0.95\columnwidth]{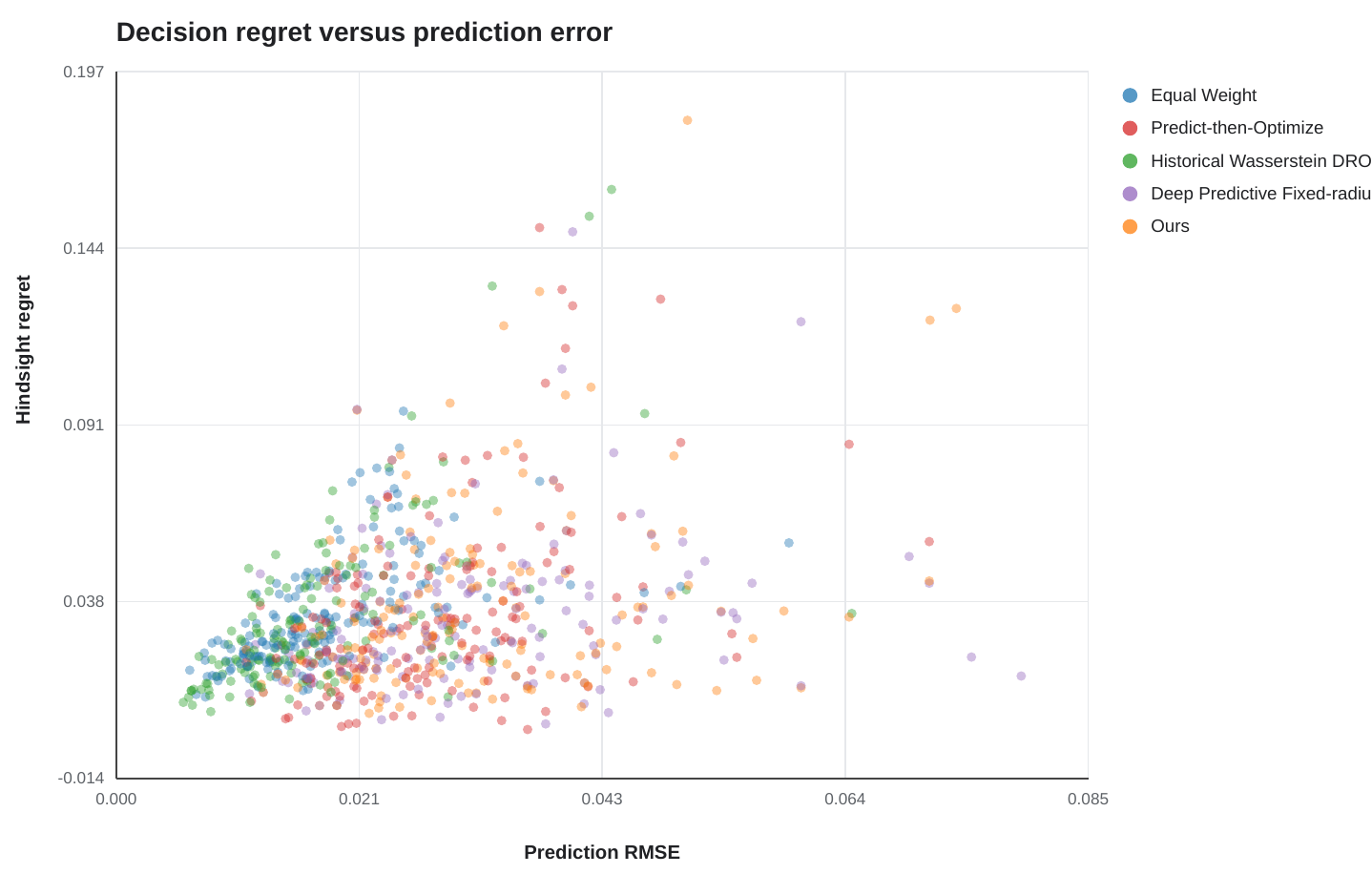}
\caption{Decision regret versus prediction error. Prediction error alone does not determine downstream loss because optimization amplifies errors differently across directions.}
\label{fig:regret_prediction}
\end{figure}

Figure~\ref{fig:regret_prediction} plots decision regret against prediction error. The diagnostic illustrates why prediction error alone is not enough: decision loss depends on whether forecast error lies in a decision-sensitive direction. This supports the decision-focused design of the learned ambiguity radius.

\section{Discussion and Limitations}

The experiments show a clear hierarchy. Nonrobust predict-then-optimize performs poorly because it turns noisy return forecasts into aggressive portfolios. Historical Wasserstein DRO is stable but underuses contextual information. Deep fixed-radius DRO is a strong baseline and achieves the highest overall Sharpe ratio in the current portfolio experiment. LPAS-W does not dominate this fixed-radius baseline on every average metric, but it obtains competitive returns with a much smaller average radius, slightly better worst-month and CVaR95 loss, and stronger adaptivity in high-radius and drawdown regimes.

The current implementation has three limitations. First, the experiment is a single rolling split over one 20-asset S\&P 500 universe; broader universes, multiple seeds, and additional rolling folds should be included for stronger empirical claims. Second, the learned anisotropic metric is part of the framework but not evaluated in the present experiment. Third, empirical coverage is below the nominal quantile level after decision-aware tuning, so strict risk-control applications should add conformal post-calibration.

\section{Conclusion}

We proposed learned predictive ambiguity sets for decision-focused DRO. The method learns a finite nominal scenario distribution and a context-dependent Wasserstein radius, then solves a robust decision layer whose ambiguity set adapts to the current state. The portfolio experiment shows that this approach substantially outperforms nonrobust and historical robust baselines, and remains competitive with a strong deep fixed-radius DRO model while using a smaller radius. The results suggest that the right object to learn in robust decision-making is not only a prediction, but also how much the optimizer should distrust that prediction.

\end{document}